\documentclass[letterpaper]{article} 
\usepackage{aaai2026}  
\usepackage{times}  
\usepackage{helvet}  
\usepackage{courier}  
\usepackage[hyphens]{url}  
\usepackage{graphicx} 
\usepackage{natbib}  
\usepackage{caption} 
\usepackage{algorithm}
\usepackage{algorithmic}

\usepackage{newfloat}
\usepackage{listings}
\DeclareCaptionStyle{ruled}{labelfont=normalfont,labelsep=colon,strut=off} 
\floatstyle{ruled}
\newfloat{listing}{tb}{lst}{}
\floatname{listing}{Listing}
\newcommand{\ours}{\textsc{FusionDP}}

\usepackage{amsmath}
\usepackage{amsthm}

\newtheorem{definition}{Definition}
\newtheorem{theorem}{Theorem}

\title{FusionDP: Foundation Model-Assisted Differentially Private Learning for Partially Sensitive Features}
\author {
    Linghui Zeng\textsuperscript{\rm 1},
    Ruixuan Liu\textsuperscript{\rm 1},
    Atiquer Rahman Sarkar\textsuperscript{\rm 2},
    Xiaoqian Jiang\textsuperscript{\rm 3},
    Joyce C. Ho\textsuperscript{\rm 1},
    Li Xiong\textsuperscript{\rm 1}
}
\affiliations {
    \textsuperscript{\rm 1}Emory University\\
    \textsuperscript{\rm 2}University of Manitoba\\
    \textsuperscript{\rm 3} University of Texas Health Science Center\\
    linghui.zeng@emory.edu, ruixuan.liu2@emory.edu, sarkarar@myumanitoba.ca, Xiaoqian.Jiang@uth.tmc.edu, joyce.c.ho@emory.edu, lxiong@emory.edu
}
\usepackage{bibentry}

\begin{document}

\maketitle

\begin{abstract}
Ensuring the privacy of sensitive training data is crucial in privacy-preserving machine learning. 
However, in practical scenarios, privacy protection may be required for only a subset of features. For instance, in ICU data, demographic attributes like age and gender pose higher privacy risks due to their re-identification potential, whereas raw lab results are generally less sensitive. Traditional DP-SGD enforces privacy protection on all features in one sample, leading to excessive noise injection and significant utility degradation.
We propose \ours, a two-step framework that enhances model utility under feature-level differential privacy. First, \ours~ leverages large foundation models to impute sensitive features given non-sensitive features, treating them as external priors that provide high-quality estimates of sensitive attributes without accessing the true values during model training. Second, we introduce a modified DP-SGD algorithm that trains models on both original and imputed features while formally preserving the privacy of the original sensitive features.
We evaluate \ours~ on two modalities: a sepsis prediction task on tabular data from PhysioNet and a clinical note classification task from MIMIC-III. 
By comparing against privacy-preserving baselines, 
our results show that \ours~ significantly improves model performance while maintaining rigorous feature-level privacy, demonstrating the potential of foundation model-driven imputation to enhance the privacy-utility trade-off for various modalities.
\end{abstract}

\section{Introduction}
Differential privacy (DP) \cite{DP} is the gold standard for protecting individual data in machine learning. Differentially Private Stochastic Gradient Descent (DP-SGD) is the most widely adopted mechanism, providing sample-level privacy by clipping gradients and injecting noise during training \cite{dpsgd}. However, DP-SGD applies protection across all features in each training sample, which can lead to excessive noise injection and significant utility degradation, especially for high-dimensional data.
In many real-world applications, not all features of training data are considered sensitive, either due to regulations or individual preferences. For instance, in clinical domains, under HIPAA \cite{hipaa1996}, only 18 specific identifiers (e.g., geographic info, dates, facial images, unique characteristics) are classified as protected health information (PHI), while other clinical features may not require the same level of protection. 
Moreover, studies have shown that providing patients with fine-grained
control over their data - including customizing privacy preferences across different data categories - is instrumental in
maintaining engagement and participation \cite{kim2017iconcur}.
Selective feature protection is needed across data modalities beyond tabular data. In natural language processing, privacy concerns focus on specific elements such as names or locations; in imaging, facial features or identifying markers may need to be obscured. Applying DP-SGD across all features introduces unnecessary noise, resulting in degraded utility, especially when only a portion of the features are identified as sensitive.

A recent work defines Feature Differential Privacy (Feature-DP)~\cite{featuredp} for the above partially sensitive scenarios and proposes mechanisms that aim to only protect the sensitive (or private) subset of features. This relaxation of the DP guarantee offers the potential for improved utility by avoiding unnecessary noise on public (or non-private) features. However, they leverage the non-sensitive features by directly masking or omitting private features during training, which can lead to significant information loss. 
In domains such as healthcare, sensitive attributes (e.g., age, intensive care unit (ICU) length of stay, or specific tokens in clinical notes) are often highly informative for prediction. As demonstrated in our experiments, removing or masking these features can significantly degrade model utility.



Given the strong prior knowledge embedded in large foundation models, it has become feasible to improve the utility by imputing sensitive features with an external foundation model. 
These models, trained on large public corpora, have the potential to generate (impute) synthetic approximations of private attributes in a reasonable range, which provides a better prior than masking dummy values. 
It should be noted that the prior is ``free" to access since foundation models are already open-sourced.
Such imputed data can serve as a “free” proxy to the sensitive features, improving model accuracy while maintaining formal privacy guarantees on the sensitive attributes.

We propose \textbf{\ours}, a two-step framework designed to improve the privacy-utility trade-off under feature-DP. First, we leverage foundation models to impute the sensitive features, forming a hybrid dataset that combines privacy-free estimates of sensitive attributes with the original public features. Second, we introduce a modified DP-SGD algorithm that formally protects only the original private features during training, while leveraging the hybrid data to guide the optimization. In particular, we incorporate a private embedding alignment objective that encourages consistency between the representations of original and hybrid inputs. This combination provides a more informative training signal that better preserves the utility of the original data, even under strong privacy constraints.

We evaluate {\ours} on two modalities: a sepsis prediction task on tabular data and a clinical note classification task, demonstrating the first application of featureDP to textual data. Our results show that {\ours} significantly improves performance across a range of privacy budgets and generalizes across modalities. By bridging feature-DP with foundation model priors, {\ours} demonstrates the potential of hybrid learning to unlock better utility while maintaining formal privacy protections on sensitive features.

\section{Related Work}


\textbf{DP-SGD} is a widely used technique to enforce sample-level differential privacy during model training \cite{dpsgd}. It operates by clipping the per-sample gradients to a fixed norm bound and injecting Gaussian noise before averaging the gradients to update the model. Formally, this mechanism ensures that the model’s output does not significantly change when any single data point is modified or removed, providing strong privacy guarantees quantified by 
$(\epsilon, \delta)$-differential privacy.

However, DP-SGD treats all features of each sample uniformly as private, resulting in excessive noise injection and substantial utility loss when only a subset of features require protection. This over-protection makes it suboptimal for practical use cases like clinical prediction, where many raw features (such as lab results) are not individually sensitive but still suffer from global noise added during training.

There are previous works that leverage publicly available dataset in DP-SGD to improve the model performance, such as via linear combination of public and private gradients~\cite{amid2022public,liu2023coupling}.
Different from ours, they are originally proposed for satisfying the conventional sample-level DP, and assume an available public dataset, while the domain-shift between public and private training data may ruin the performance.




\textbf{Feature-DP} protects only a subset of features (e.g., sensitive attributes like race or education level) during training \cite{featuredp}. This approach formalizes feature-level privacy as a relaxation of traditional DP, ensuring that an adversary cannot infer the presence of a data point with respect to the protected features, even when the public features are known.
To achieve this, \citet{featuredp} propose a variant of DP-SGD tailored to feature-DP, which separates the loss into a public loss (on public features) and a private loss (involving private features). Privacy amplification is achieved by sampling two separate batches for each loss in each training iteration. \ours~also aims to achieve feature-DP and targets to further improve the utility.

\section{Problem Setting}

We consider a scenario where only a subset of features requires privacy protection, while the rest can be used without restrictions. 
Formally, let $\mathcal{X}$ denote the data space, where each data point $x \in \mathcal{X}$ can be decomposed into sensitive (private) components $x_{\text{priv}}$ and non-sensitive (public) components $x_{\text{pub}}$, such that $x = (x_{\text{priv}}, x_{\text{pub}})$. 
This setting commonly arises in real-world applications such as healthcare, finance, and social platforms, where $x_{\text{priv}}$ includes features like demographic attributes, rare diagnoses, education history, or identifiable spans in text that pose greater re-identification risks, or sensitive features specified by users. Meanwhile, $x_{\text{pub}}$ encompasses features like lab results, transactions, or non-identifying text tokens that do not require the same level of protection.

The goal is to train a model $f: \mathcal{X} \to \mathcal{Y}$ that achieves high utility while satisfying DP constraints only with respect to the sensitive features.
To perform this selective protection, we adopt the notion of \emph{Feature-DP}~\cite{featuredp}, which generalizes traditional sample-level DP to protect only designated features while allowing unrestricted use of public features during training.

\begin{definition}[Feature Differential Privacy~\cite{featuredp}]
Let $\Psi : \mathcal{X} \to \mathcal{F}$ be a projection that maps an input $x \in \mathcal{X}$ to its public features $\Psi(x)$, and let $f : [0,1] \to [0,1]$ be a trade-off function. A randomized mechanism $\mathcal{M}$ is $f$-feature differentially private with replacement, relative to $\Psi$, denoted $f$-DP$^\Psi_r$, if for all neighboring datasets $S \triangle S' = \{x, x'\}$ such that $\Psi(x) = \Psi(x')$, and for all measurable subsets $\mathcal{T}$ of the output space:
\[
\Pr[\mathcal{M}(S) \in \mathcal{T}] \leq 1 - f\big(\Pr[\mathcal{M}(S') \in \mathcal{T}]\big).
\]

Moreover, we say a mechanism is $(\epsilon, \delta)$-$\mathrm{DP}^{\Psi}_r$ iff it is $f$-$\mathrm{DP}^{\Psi}_r$ for $f(x) = 1 - \delta - e^{\epsilon} \cdot x$.
\end{definition}






Our goal is to design a training algorithm that satisfies the feature-DP definition while achieving high utility on downstream tasks. In particular, we address the challenge of information loss caused by either removing sensitive features or injecting excessive noise into gradients. To overcome this, we leverage foundation models with publicly available priors to obtain high-quality estimates of sensitive features while breaking their direct link to individual identity, thereby improving model utility under feature-DP constraints. It's important to note that, the foundation models are pre-trained on public data only and has no access to the true sensitive features $x_{priv}$ from our dataset during training or inference.

\section{\ours}

\begin{figure}[t]
\centering
\includegraphics[width=1.0\columnwidth]{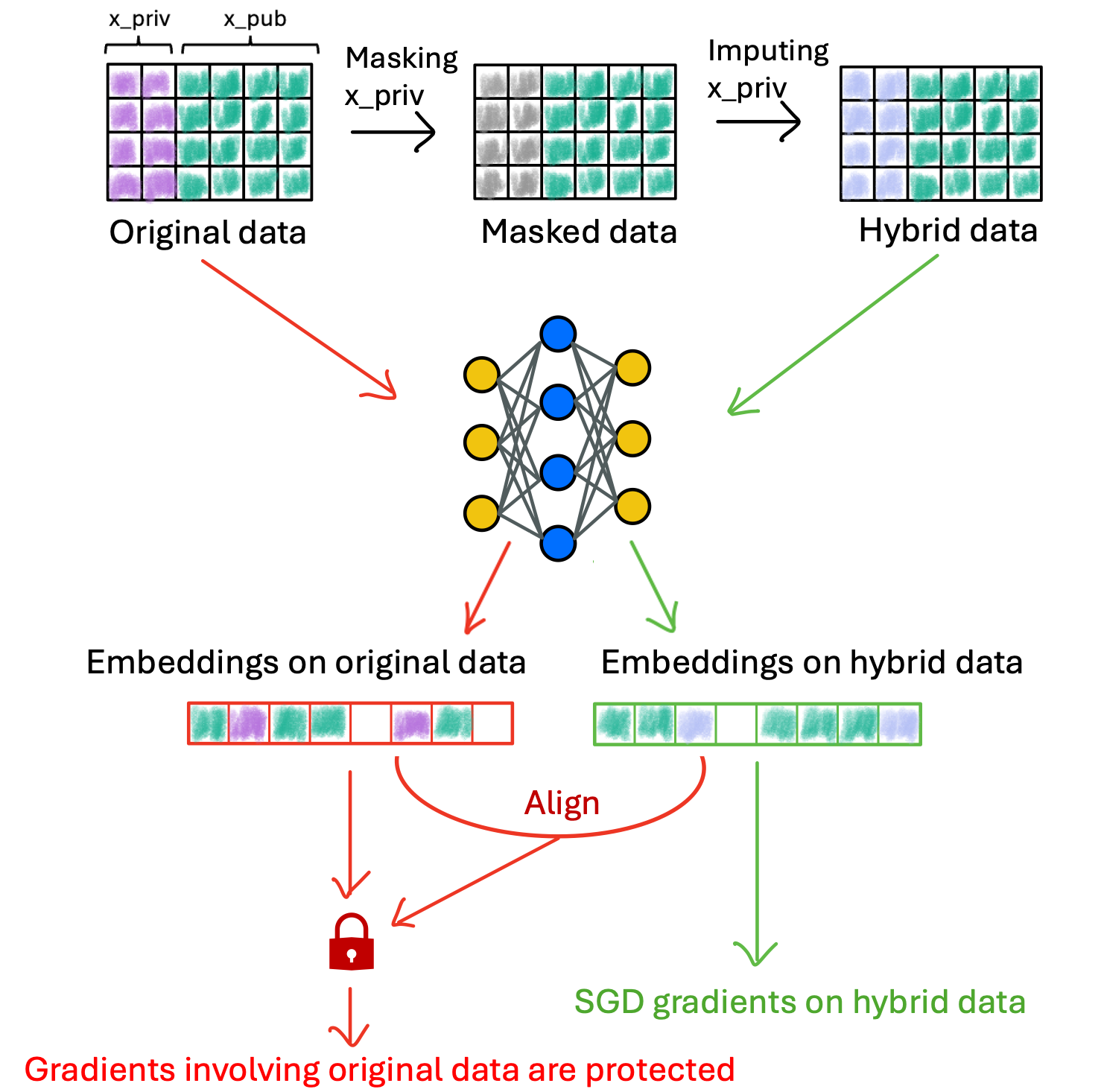} 
\caption{Overview of \ours. Sensitive (purple) features in the original data are masked and then imputed (space gray) using foundation models. The hybrid data is used for public gradient updates. Gradients involving private data (red arrows) are protected by clipping and noises.}
\label{fig:fusiondp}
\end{figure}

\ours~is a two-step framework to achieve feature-DP with improved utility. Figure \ref{fig:fusiondp} illustrates this training pipeline. We first impute sensitive features using a foundation model, producing a hybrid sample in which private attributes are replaced by high-quality estimates based on the non-sensitive features. We then train the model with a combined loss objective of public (in green) and private (in red) components.
We clip and add noise only to the gradient of the private loss, which isolates and bounds the contribution of private features.
Under this framework, we improve the private gradient component by 
leveraging the gradient calibration and proposing a representation-consistency regularizer to align hidden states of original and hybrid inputs. 

\subsection{Imputation with Foundation Models}

To preserve privacy while improving utility, we leverage foundation models to create the hybrid training data.
We use a pretrained foundation model $\mathcal{F}$ (e.g., TabPFN \cite{tabpfn} for tabular data, and GPT-4o mini for text) to predict the masked sensitive features conditioned on the public features:
\[
\hat{x}_{\text{priv}} = \mathcal{F}(x_{\text{pub}}).
\]

Crucially, $\mathcal{F}$ is a fixed, pre-trained and publicly available model that was trained without access to any sensitive features from our private dataset. 
And $\hat{x}_{\text{priv}}$ is generated without observing $x_{\text{priv}}$.
Thus, the imputation process itself does not leak information about the true $x_{priv}$ values, and we treat the imputed data as public.
The resulting \emph{hybrid} sample is
\[
\tilde{x} = (x_{\text{pub}}, \hat{x}_{\text{priv}}).
\]

Previous public-assisted works \cite{amid2022public,liu2023coupling} typically assume access to an external public dataset.
However, when the public data has shifted distribution from the private dataset, the divergent gradients result in low utility. In contrast, our approach leverages foundation models to impute sensitive features, synthesizing hybrid data that better preserves the distributional characteristics of the original dataset. \ours~then benefits from cleaner, task-relevant public gradient, instead of relying on the availability of an external public dataset with potential distribution shift.

This approach further enables us to split training into two branches: DP-SGD is applied to gradients involving original data with formal privacy guarantees, while standard SGD is used on hybrid data. DP-SGD clips and perturbs gradients but the sample-level protection can obscure useful signals, especially when the public features are helpful for the task
In contrast, SGD on hybrid data provides a cleaner training signal without perturbation, true public values and high-quality priors of sensitive features, though at the risk of imputation bias. 
The imputation details can be found in Section \ref{sec:experiments}.

\subsection{Model Training}
\label{sec:training}


We formulate a general \textbf{hybrid training framework} for feature DP. Inspired by public-assisted DP-SGD \cite{amid2022public, liu2023coupling}, the idea is to linearly combine private and public gradient :  
\[
g_t \leftarrow g^{\text{pub}}_t + \alpha\, \tilde g^{\text{priv}}_t
\]
Here, $g^{\text{pub}}$ denotes the public gradient computed on hybrid data, which includes imputed sensitive features and does not access their true values: $g_{pub} = \nabla_\theta \ell_{pub} = \nabla_\theta(\ell(f_\theta(\tilde{x}), y))$.

In contrast, $g^{\text{priv}}$ is the private gradient derived from a loss term (as we will define next) on a batch of original data that includes sensitive features.
The per-sample gradients in the batch are processed  via gradient clipping and perturbation in DP-SGD to ensure feature-DP protection:
\[
\tilde g^{\text{priv}}_t = \text{DP-SGD}(\nabla_\theta \ell_\text{priv}),
\]

Under this framework, we instantiate multiple variants of the \emph{private} loss $\ell_{\text{priv}}$ which progressively improve privacy-utility trade-offs.

\begin{itemize}
    \item \textbf{Na\"ive Fusion}: $\ell_{\text{priv}} = \ell(f_\theta(x), y)$  
    The private gradient is directly computed on the standard loss of the original data with sensitive features. 
    Compared to public-assisted DP-SGD methods~\cite{amid2022public, liu2023coupling} that rely on separate public datasets and may suffer from utility drop if the public data distribution shifts from the original sensitive data, our hybrid data—where sensitive features are imputed by foundation models—better matches the original distribution, especially when the number of sensitive features is small, providing a better prior and leading to improved performance.
    
    \item \textbf{Calibrated Fusion}: $\ell_{\text{priv}} = \ell(f_\theta(x), y) -  \ell(f_\theta(\tilde{x}), y)$  
    This variant isolates the influence of sensitive features by calibrating or subtracting the loss on original data with the loss on hybrid data, 
    which follows the similar form as previous works~\cite{DOPE, featuredp}, but has the key difference that our public component is obtained on hybrid data where sensitive features are imputed.
    Because the gradient on the calibrated loss has a lower norm compared to $\nabla_\theta (\ell(f_\theta(x), y))$~\cite{DOPE}, the noise magnitude can be reduced due to a lower sensitivity.

    \item \textbf{\ours~ (Ours)}:  
    \begin{align}
    \ell_{\text{priv}} &=  \ell(f_\theta(x), y) - \ell(f_\theta(\tilde{x}), y)
    + \\
    &\beta \|h_\theta(x) - h_\theta(\tilde{x})\|_2^2
    \end{align}
    where $h_\theta(\cdot)$ denotes the hidden representation of the input (e.g., the output of an intermediate layer before classification), $\beta$ denotes the regularization coefficient that controls the strength of the representation consistency term. In addition to the calibrated loss, we are inspired by previous representation learning works~\cite{sajjadi2016regularizationstochastictransformationsperturbations, chen2020simpleframeworkcontrastivelearning} and introduce the \emph{Representation Consistency} regularizer that aligns hidden representations of $x$ and $\tilde{x}$ by minimizing the distance between them.  
    This encourages the model to treat original and hybrid inputs similarly, mitigating bias from imperfect imputation, helping the $g_\text{pub}$ benefit indirectly from $\tilde{g}_\text{priv}$ and further stabilizing training. 
\end{itemize}



We use two independent mini-batches in each iteration to enable privacy amplification by sub-sampling, as proved by \cite{featuredp}:
(i) a Poisson-sampled batch $B_{\text{priv}}$ for the private term, and  
(ii) a uniformly sampled batch $B_{\text{pub}}$ (typically larger) for the public term.  
Only the per-sample gradients computed with original data ($g^{\text{priv}}_i$) are clipped and perturbed; gradients from public batch using only the hybrid data ($g^{\text{pub}}_t$) are added without noise.
The complete training procedures of \ours~is in Algorithm \ref{alg:fusiondp}.

\begin{algorithm}[t]
\caption{FusionDP Training Procedure}
\label{alg:fusiondp}
\footnotesize
\begin{algorithmic}[1]
\REQUIRE Dataset $\mathcal{D}=\{(x_i,y_i)\}_{i=1}^N$, where $x_i=(x_{\text{pub}},x_{\text{priv}})$; foundation model $\mathcal{F}$; clipping norm $C$; noise multiplier $\sigma$; learning rate schedule $\eta_t$; consistency weight $\beta$; mixing weight $\alpha$
\ENSURE Trained parameters $\theta_T$ of model $f_\theta$

\STATE Initialize $\theta_0$ randomly.

\FOR{$t=0$ \TO $T-1$}
    \STATE \textbf{Sample batches:}
    \STATE \hspace{0.5em} Private batch $B^{\text{priv}}_t$ via Poisson sampling with prob.\ $p$.
    \STATE \hspace{0.5em} Public batch $B^{\text{pub}}_t$ uniformly at random (size $m'$).

    \STATE \textbf{Public branch (clean SGD):}
    \FORALL{$(x,y)\in B^{\text{pub}}_t$}
        \STATE $\hat{x}_{\text{priv}} \leftarrow \mathcal{F}(x_{\text{pub}})$; \quad $\tilde{x} \leftarrow (x_{\text{pub}}, \hat{x}_{\text{priv}})$
        \STATE $\ell_{\text{pub}}(x,y) \leftarrow \ell\!\big(f_\theta(\tilde{x}),y\big)$
        \STATE $g^{\text{pub}}_i \leftarrow \nabla_\theta \ell_{\text{pub}}(x,y)$
    \ENDFOR
    \STATE $g^{\text{pub}}_t \leftarrow \frac{1}{|B^{\text{pub}}_t|}\sum_{i\in B^{\text{pub}}_t} g^{\text{pub}}_i$

    \STATE \textbf{Private branch (DP-SGD):}
    \FORALL{$(x,y)\in B^{\text{priv}}_t$}
        \STATE $\hat{x}_{\text{priv}} \leftarrow \mathcal{F}(x_{\text{pub}})$; \quad $\tilde{x} \leftarrow (x_{\text{pub}}, \hat{x}_{\text{priv}})$
        \STATE $\ell_{\text{priv}}(x,y) \leftarrow \ell\!\big(f_\theta(x),y\big) - \ell\!\big(f_\theta(\tilde{x}),y\big) + \beta \,\|h_\theta(x)-h_\theta(\tilde{x})\|_2^2$
        \STATE $g^{\text{priv}}_i \leftarrow \nabla_\theta \ell_{\text{priv}}(x,y)$
        \STATE \textbf{Clip: } $\bar g_i \leftarrow g^{\text{priv}}_i / \max\!\big(1,\|g^{\text{priv}}_i\|_2/C\big)$
    \ENDFOR
    \STATE $\tilde g^{\text{priv}}_t \leftarrow \frac{1}{|B^{\text{priv}}_t|}\left(\sum_{i\in B^{\text{priv}}_t}\bar g_i \;+\; \mathcal{N}(0,\sigma^2 C^2 \mathbf{I})\right)$

    \STATE \textbf{Combine and update:}
    \STATE $g_t \leftarrow g^{\text{pub}}_t + \alpha\, \tilde g^{\text{priv}}_t$
    \STATE $\theta_{t+1} \leftarrow \theta_t - \eta_t\, g_t$
\ENDFOR
\RETURN $\theta_T$
\end{algorithmic}
\end{algorithm}

\subsection{Privacy Analysis}

We follow the feature-DP accounting of \citet{featuredp}.  
Only the gradient of \emph{private} loss is clipped and perturbed; the public branch (on hybrid data) does not touch any sensitive values and does not affect privacy.



\begin{theorem}[\ours~ Privacy, $(\varepsilon,\delta)$-\textnormal{DP}$^{\psi}_i$]\label{thm:fusiondp}
Let the per-example gradient of the private loss be clipped to norm $C$, and let $p$ be the Poisson sampling rate for the private batch. Suppose the private loss is computed using the original input $x = (x_{\text{pub}}, x_{\text{priv}})$ and the imputed hybrid input $\tilde{x} = (x_{\text{pub}}, \hat{x}_{\text{priv}})$, where its public features $x_{\text{pub}} = \Psi(x)$ are extracted via a public feature extractor $\Psi$, and $\hat{x}_{\text{priv}} = \mathcal{F}(x_{\text{pub}})$ is generated by a fixed foundation model $\mathcal{F}$ that has been pre-trained on public data without access to the private dataset $\mathcal{D}$. Then after $T$ steps, \textsc{FusionDP} satisfies
\[
f\textnormal{-DP}^{\Psi}_i, \quad f = T\!\left( \mathcal{N}(0, \sigma)^T,\; \left((1-p)\mathcal{N}(0, \sigma) + p \, \mathcal{N}(\tau, \sigma)\right)^T \right),
\]
where $\tau$ is the (post-clipping) Lipschitz bound of the private loss, and $\sigma$ is the Gaussian noise scale.

In particular, choosing

\begin{align}
\label{sigma}
\sigma = \frac{c\,\tau\,m}{\varepsilon\,n} \sqrt{T \log \frac{1}{\delta} \log \frac{T}{\delta}},
\end{align}

for absolute constant $c$, private batch size $m$, and dataset size $n$, yields an $(\varepsilon, \delta)$-\textnormal{DP}$^{\Psi}_i$ guarantee with respect to the private features $x_{\text{priv}}$.
\end{theorem}

\noindent\textbf{Why the theorem holds.} The statement is a direct corollary of Theorem~5 in \citet{featuredp}:  
(i) the public gradient uses only imputed $\hat{x}_{\text{priv}}$ and therefore does not increase privacy cost;  
(ii) the representation consistency loss operates on the original and hybrid inputs but is encapsulated within the private loss and bounded by clipping and noise; 
(iii) most importantly, the use of two independently sampled mini-batches—one Poisson-sampled for the private gradient and one uniform for the public gradient—enables valid privacy amplification by sub-sampling, a critical assumption in the feature-DP framework.  

\paragraph{Remark.} 
All variants of private loss within our hybrid training framework maintain the $f\textnormal{-DP}^{\Psi}_i$, as long as the influence of sensitive features in each sample remains bounded when gradients are appropriately privatized.
Following feature-DP~\cite{featuredp}, the theorem is based on the Lipschitz constant, while the privacy guarantee holds when we bound the sensitivity via gradient clipping instead of the Lipschitz constant.


\section{Experiments}
\label{sec:experiments}

In this section, we demonstrate the effectiveness of the proposed approach through comprehensive evaluations on two distinct modalities: a  prediction task on tabular data from the PhysioNet Sepsis Prediction Challenge and a clinical text classification task using MIMIC-III. We compare \ours{} against a suite of baseline methods and ablation variants to validate its performance and robustness.

\subsection{Baseline Methods}

We compare \ours~with the three non-DP baselines, three DP approaches, and 2 feature-DP variants.
\begin{enumerate}
    \item \textsc{SGD-ORG}: A non-DP model trained using SGD on the original data $x$, serving as the utility upper bound.
    \item \textsc{SGD-Hybrid}: A non-DP model trained on the hybrid dataset, $x_{\text{hybrid}}$ where sensitive features have been imputed using foundation models.
    \item \textsc{SGD-Pub}: A non-DP model trained on a public dataset with sensitive features masked using Gaussian noise.
    \item \textsc{DP-SGD} \cite{dpsgd}: A standard DP-SGD model that enforces sample-level differential privacy, protecting all features through clipping and noise addition during training.
    \item \textsc{Na\"ive Fusion}: A na\"ive DP baseline combining gradients from \textsc{DP-SGD} and \textsc{SGD-Hybrid} using a weighted average.
    \item \textsc{Na\"ive Fusion-Pub}: A na\"ive DP baseline combining gradients from \textsc{DP-SGD} and \textsc{SGD-Pub} using a weighted average.
    \item \textsc{Feature-DP}: The original feature-DP algorithm used in \citet{featuredp}.
    \item \textsc{Calibrated Fusion}: An ablation of \ours~without representation consistency regularization, isolating its contribution to performance. While both 
    \textsc{Calibrated Fusion} and \textsc{Feature-DP} use a calibrated loss formulation, 
    \textsc{Feature-DP} masks sensitive features whereas \textsc{Calibrated Fusion} 
    uses imputed sensitive features in the hybrid data.
\end{enumerate}
Additional baseline details are available in the Appendix.

\subsection{Tabular Sepsis Prediction}

We begin by evaluating \ours~on a tabular classification task involving early prediction of sepsis.

\subsubsection{Dataset.}

We use the PhysioNet 2019 Challenge dataset \cite{physionet, sepsis1, sepsis2}, a large-scale tabular ICU dataset comprising over 40{,}000 patient records. Each record includes time-series and static features such as vital signs, laboratory test results, and demographics, along with binary labels indicating the onset of sepsis. 

In our experimental setting, we designate all the demographic attributes (i.e., age, gender, ICU unit type, the hours between hospital and ICU admission, and ICU length of stay) as \emph{private features}, due to their heightened risk of re-identification. The remaining features, including clinical and physiological measurements, are considered public. This setup allows us to evaluate models under a realistic and practical partial-feature privacy setting.

\subsubsection{Data Imputation.}
\label{sec:tabular-imputation}
We use TabPFN~\cite{tabpfn}, a transformer-based tabular foundation model pretrained on millions of synthetic tabular datasets, to impute sensitive features. TabPFN operates via meta-learning, using a support set to make Bayesian posterior predictions on a query set in a single forward pass. Although LLMs can potentially be used for imputation, they are less suited to structured tabular data and require extensive post-processing to convert free-text responses into usable feature formats. TabPFN natively supports structured data with reliable outputs, making it more practical for tabular feature imputation.  We treat non-sensitive features as input features and the sensitive features as TabPFN target labels for the support set and impute sensitive demographic attributes for query samples, substituting private values with high-quality estimates.

The dataset is partitioned as 10\% public samples (TabPFN support set), 70\% query set (training data for \ours), and 20\% split equally for validation and test sets. Note that, the 10\% support set is reasonably small and practical, and can represent data from consented patients who agreed to broader data sharing, a common scenario in clinical settings where some patients opt into research studies with less restrictive privacy requirements.
For imputation, we condition TabPFN on the public support set and generate predictions for the sensitive features in the query set. Each sensitive attribute is treated as the target label in a TabPFN classifier or regressor, depending on its type, and the non-sensitive features serve as input. We then combine the imputed sensitive features with the original non-sensitive features to construct the hybrid dataset.

\subsubsection{Backbone model.}
For these experiments, we use a lightweight multi-layer perceptron (MLP) as the backbone model. Additional data preprocessing and backbone information are in the Appendix. 

\subsubsection{Evaluation Metrics.}
The Sepsis dataset presents a binary classification task with substantial class imbalance, with only 15\% of samples labeled positive. For this imbalanced setting, we use Area Under the Precision-Recall Curve (AUPRC) rather than Area Under the ROC Curve (AUROC) as our primary metric, since AUPRC better emphasizes the model's ability to correctly identify the minority class by capturing the precision-recall trade-off. AUPRC is particularly challenging to improve in imbalanced regimes, as gains require reducing false positives while capturing true positives in a sparse positive space. A random classifier achieves AUPRC equal to the positive class prevalence ($\sim$0.15), making even modest improvements meaningful.

\subsubsection{Results.}

\begin{figure}[t]
\centering
\includegraphics[width=0.99\linewidth]{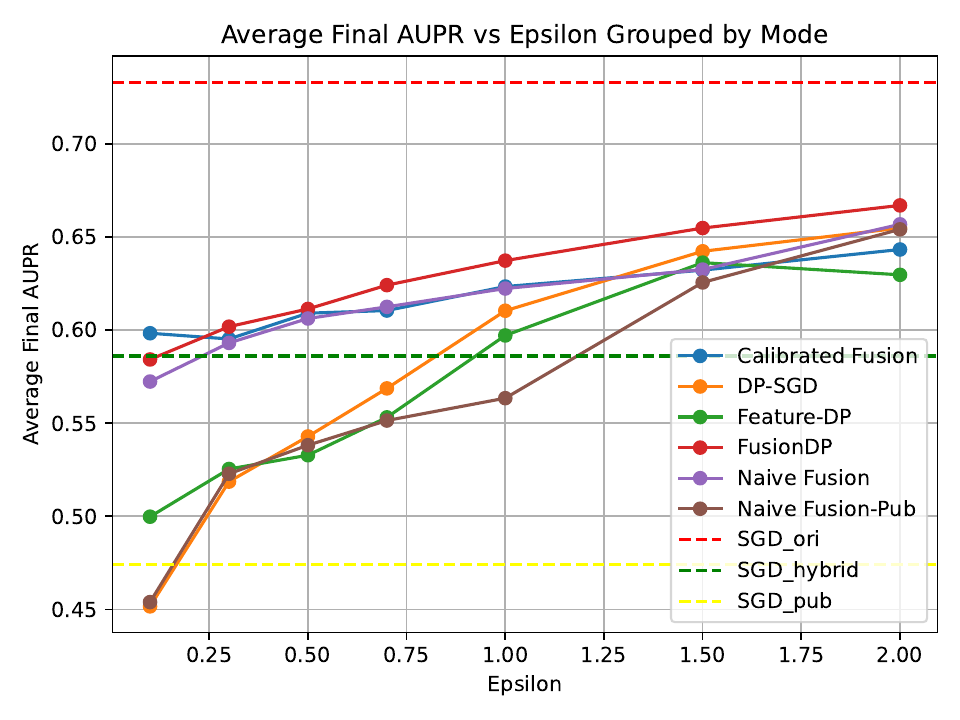} 
\caption{Best AUPRC (average of 3 runs) on the Sepsis prediction task across privacy budgets ($\epsilon$).}
\label{sepsis_aupr}
\end{figure}

Figure \ref{sepsis_aupr} reports the best AUPRC for each method across a range of privacy budgets ($\epsilon$) on the Sepsis prediction task, on average of 3 runs. For each $\epsilon$, we perform a grid search over hyperparameters using the validation set, and report the model performance on the test set. Detailed hyperparameter analysis can be found in Appendix.

Non-private baselines and standard DP establish the expected performance bounds. The substantial gap between \textsc{SGD-Hybrid} (trained on hybrid data) and \textsc{SGD-Pub} (trained on public data with masked sensitive features) demonstrates that TabPFN imputation preserves critical predictive information compared to simple masking. \textsc{DP-SGD} exhibits steep performance degradation under strong privacy constraints ($\epsilon = 0.1$) and consistently underperforms \ours{} due to noise injection across all features, including non-sensitive ones, confirming that indiscriminate privacy protection significantly hurts utility in high-privacy regimes.



\textsc{Na\"ive Fusion} yields modest improvements under strong privacy constraints (low $\epsilon$) but diminishing returns as $\epsilon$ increases. 
\textsc{Feature-DP} decomposes the loss into public and private components, reducing sensitivity of private gradients and enabling smaller noise injection. However, masking sensitive features hurts its utility, as evidenced by poor \textsc{SGD-Pub} and \textsc{Na\"ive Fusion-Pub} performance, with effectiveness depending critically on the distribution shift between public and private dataset.

\ours{} achieves superior performance through two key innovations. \textsc{Calibrated Fusion} improves over \textsc{Feature-DP} by using TabPFN-imputed hybrid data instead of masked features, benefiting from both reduced private gradient sensitivity and higher-quality feature approximation. \textsc{\ours} further introduces the representation consistency loss that aligns original and hybrid data representations, achieving the best performance across all $\epsilon$ values with consistent advantages over all baselines.

\subsection{Clinical Notes Classification}

\begin{figure}[t]
\centering
\includegraphics[width=0.8\linewidth]{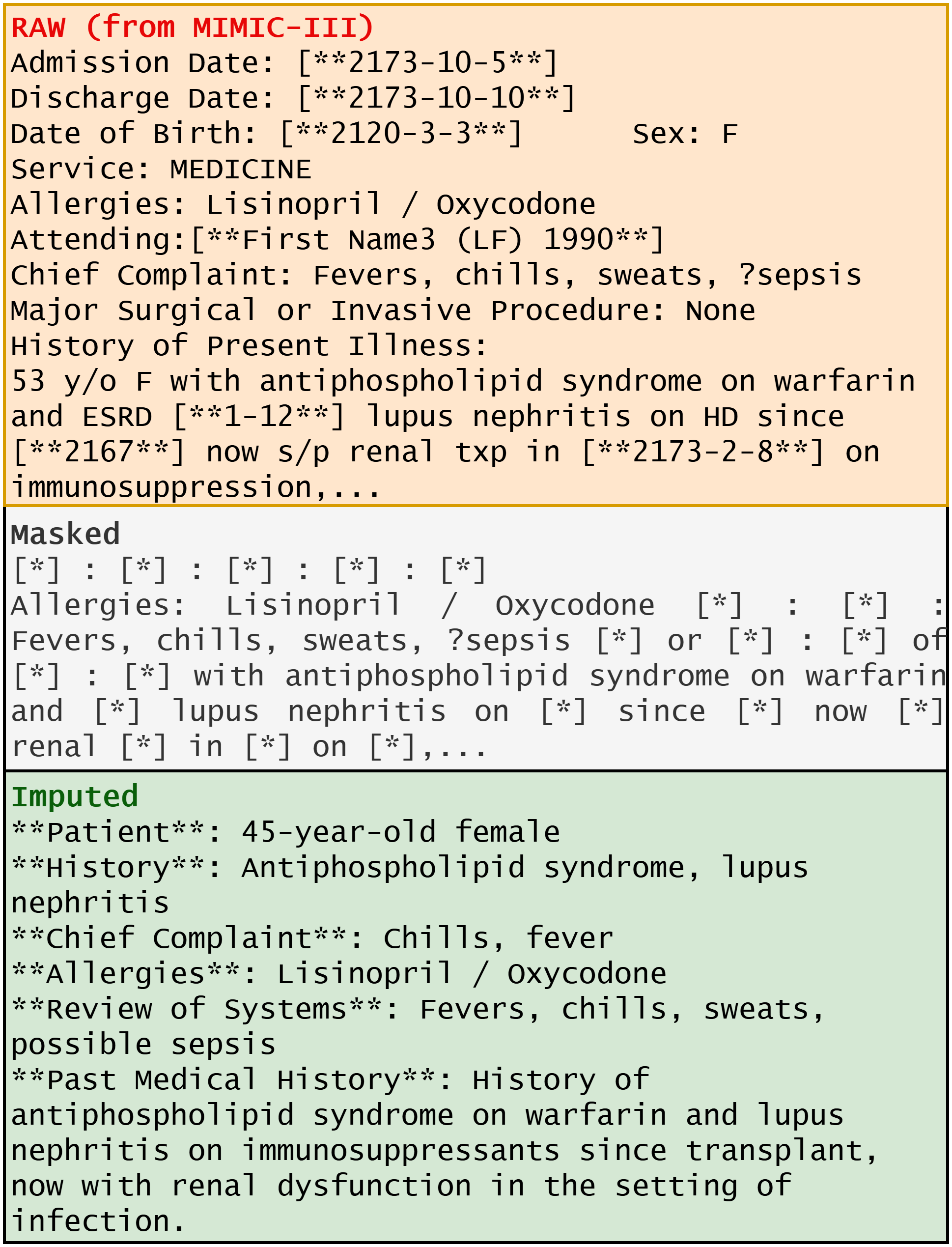} 
\caption{Snippet illustrating the flow from raw (original) clinical text to masked (public) text to imputed (hybrid) text.}
\label{fig-impute}
\end{figure}

We then evaluate \ours~on the text modality using a clinical notes classification task, to demonstrate generalizability across different modalities.

\subsubsection{Dataset.}

We use clinical notes from the MIMIC-III Clinical Database \cite{mimic3-1, mimic3-2}, a publicly available dataset comprising deidentified electronic health records of over 40,000 ICU patients from the Beth Israel Deaconess Medical Center between 2001 and 2012. The dataset includes a wide range of information such as demographics, vital signs, laboratory test results, imaging reports, and free-text clinical notes.

For our experiments, we focus on the discharge summaries, which are known to be comprehensive and consistently labeled. Each note is annotated with one or more ICD-9 diagnosis codes assigned at discharge, forming a multi-label classification task.
Following prior MIMIC-III clinical notes classification work \cite{mimic-caml, laat}, we restrict the label space to the 50 most frequent ICD-9 diagnosis codes to ensure sufficient label coverage and reliable training signal. We also adopt the same preprocessing and dataset splits.

\subsubsection{Data Pre-processing and Imputation.}
\label{sec:text-imputation}

To ensure the removal of sensitive content from clinical notes, we adopt the aggressive masking strategy described as Variation 3 in the hybrid note generation framework proposed by \citet{sarkar-notfullysynth}. This strategy integrates both de-identification and the selective retention of non-sensitive terms. Specifically, entities such as persons, dates, times, geopolitical locations, and organizations identified using the Flair Named Entity Recognition (NER) tagger \cite{flair} are systematically masked. In contrast, common stop-words, biomedical terms and phrases, numerical quantities, and items from a curated safe-word list are preserved. Additionally, biomedical acronyms as well as tokens containing alphanumeric combinations (e.g., those resembling medical record or device identifiers) are masked to further mitigate re-identification risks. This layered redaction process aims to enhance privacy protection by masking elements that could directly or indirectly reveal patient identity, which we treat as private features $x_{priv}$, while the preserved content constitutes public features $x_{pub}$.

\begin{table*}[t]
\centering

\footnotesize
\begin{tabular}{l cc c cc c cc c cc}
\hline
 & 
\multicolumn{2}{c}{\textbf{F1}} & &
\multicolumn{2}{c}{\textbf{Precision}} &&
\multicolumn{2}{c}{\textbf{Recall}} &&
\multicolumn{2}{c}{\textbf{AUC}} \\
\cline{2-3} \cline{5-6} \cline{8-9} \cline{11-12}
\textbf{Model} & Micro & Macro && Micro & Macro && Micro & Macro && Micro & Macro \\
\hline
SGD-ORG (Non-Private) & 0.4625 & 0.3427 && 0.5602 & 0.4188 && 0.3950 & 0.3129 && 0.8208 & 0.7674 \\
DP-SGD (Sample-level DP) & 0.2192 & 0.1547 && 0.2139 & 0.1593 && 0.2273 & 0.1736 && 0.6324 & 0.5513 \\
\hline
SGD-Pub & 0.2393 & 0.1913 && 0.1522 & 0.1361 && 0.5866 & 0.5218 && 0.5952 & 0.5603 \\
SGD-Hybrid & 0.3023 & 0.2114 && 0.2367 & 0.1802 && 0.4236 & 0.3254 && 0.6783 & 0.6285 \\
\hline
Na\"ive Fusion-Pub & 0.2592 & 0.1532 && 0.3874 & 0.2130 && 0.1950 & 0.1326 && 0.7032 & 0.6296 \\
Feature-DP & 0.2699 & 0.1661 && 0.3931 & 0.2347 && 0.2056 & 0.1455 && 0.7091 & 0.6574 \\
\hline
Na\"ive Fusion & 0.3064 & 0.1985 && 0.4095 & 0.2589 && 0.2461 & 0.1760 && 0.7456 & 0.6864 \\
Calibrated Fusion & 0.3220 & 0.2114 && 0.4109 & 0.2650 && 0.2666 & 0.1911 && 0.7440 & 0.6839 \\
\ours & \textbf{0.3468} & \textbf{0.2244} && \textbf{0.4854} & \textbf{0.3081} && \textbf{0.2702} & \textbf{0.1958} && \textbf{0.7666} & \textbf{0.7072} \\
\hline

\end{tabular}
\caption{Performance comparison on the MIMIC-III clinical notes classification under $\epsilon = 5.0, \delta=N^{-1.1}$, over an average of 5 independent runs; corresponding standard deviations are listed in Appendix.}
\label{tab:mimic3_metrics}
\end{table*}

The masked segments of clinical notes are imputed using GPT-4o-mini. 
The GPT-4o-mini model is accessed through the HIPAA-compliant Azure OpenAI service, in accordance with the MIMIC-III data use agreement.  We adopt the imputation prompt proposed by \citet{sarkar-notfullysynth}, which instructs the LLM to complete the masked segments in the clinical note. Figure \ref{fig-impute} presents a snippet illustrating an example. The complete prompt is provided in the Appendix. The full imputation process, conducted on the training and validation corpus comprising approximately nine thousand clinical notes, required nearly 19 hours to complete.

\subsubsection{Backbone model.}

We adopt a convolutional label-attention architecture based on MIMIC-CAML (Convolutional Attention for Multi-Label classification) \cite{mimic-caml} as our backbone model for clinical note classification. Specifically, we implement a variant of MIMIC-CAML designed to be compatible with DP-SGD training.

Unlike prior works \cite{mimic-caml, laat}, which pre-trained word2vec embeddings directly on the MIMIC-III discharge summaries, we avoid any use of the training notes to preserve privacy. Instead, we initialize the word embedding layer with publicly available word2vec vectors trained on the PMC Open Access Subset \cite{word2vec, pmc-emb, pmc_open_access}, ensuring that the model does not directly access any private clinical text during pretraining.

\subsubsection{Evaluation Metrics.}

We evaluate model performance on the MIMIC-III clinical notes classification task using four standard multilabel metrics: F1-score, Precision, Recall, and AUC (Area Under the ROC Curve). Each metric is computed with both micro and macro averaging, where micro-averaged metrics aggregate contributions of all labels across all samples to emphasize performance on frequent labels and overall classification quality, while macro-averaged metrics compute the metric independently for each label and then average them to assess the model's ability to handle rare diagnoses. 
We report test set performance using the model checkpoint that achieves the best micro-F1 score on the validation set, averaged over 5 independent runs.




\subsubsection{Results.}

Table~\ref{tab:mimic3_metrics} compares \ours~and baselines on predicting ICD-9 codes from MIMIC-III discharge summaries under $\epsilon = 5$. Following the range of previous works~\cite{yu2021differentially,de2022unlocking}, we use a moderate budget of $\epsilon = 5$, which provides a balance between meaningful privacy guarantees and model utility for clinical text classification.



\ours~achieves the best or second-best performance across all metrics, outperforming all private baselines in F1 (micro: 3468, macro: 0.2244), Precision (micro: 0.4854, macro: 0.3081), Recall (micro: 0.2702, macro: 0.1958) and AUC (micro: 0.7666, macro: 0.7072). This demonstrates strong classification performance across both frequent and rare labels. \textsc{Calibrated Fusion} shows noticeable drops across all metrics when representation consistency loss is removed, confirming this component's effectiveness. \textsc{Feature DP}, operating on redacted notes with masked sensitive spans, shows significant performance degradation compared to \textsc{Calibrated Fusion}, highlighting that simple masking removes clinically critical information. 

These results confirm that tailoring privacy protection to sensitive features, rather than indiscriminate noise injection, significantly improves utility under DP. High-quality imputation of sensitive spans proves crucial for preserving predictive performance by mitigating information loss from masking, with \ours~providing an effective framework for clinical text applications.

\section{Conclusion}

We proposed \ours, a novel framework that leverages foundation models to impute sensitive features and introduces a new training algorithm with alignment between private and imputed data that improves the privacy-utility trade-off under feature-DP. 
\ours~is evaluated on two modalities, tabular and free-text clinical data, demonstrating strong performance and extending feature-DP to natural language, which has not been previously explored.

Our current work assumes labels are public and focuses solely on sensitive feature protection. Additionally, identifying which features are sensitive is outside the scope of this paper. In practice, this determination must be made carefully by domain experts or data providers.
Future work includes extending our framework to protect labels, as well as adapting \ours~to multi-modal settings such as medical imaging paired with text reports (e.g., MRI scans with radiology captions), where selective privacy is equally important.

\bigskip

\bibliography{aaai2026}
\clearpage
\appendix




\section{Experimental Setting Details}

\subsection{Baseline Method Details}
\label{app:baseline}

We compare \ours~with the following baselines/variants:

\begin{enumerate}
    \item \textsc{SGD\_ori}: A non-private model trained on the original data using standard SGD. The model is update with:
    \[
    g = \nabla \mathcal{L}(x_{\text{ori}}, y).
    \]
    where \( x_{\text{ori}} \) is the original data. This serves as a utility upper bound, representing the best performance achievable without any privacy constraints.
    \item \texttt{SGD\_hybrid}: A non-DP model trained on a hybrid dataset, where sensitive features have been imputed using foundation models. The model is update with:
    \[
    g = \nabla \mathcal{L}(x_{\text{hybrid}}, y).
    \]
    where \( x_{\text{hybrid}} \) is the hybrid data. This is conceptually similar to training on synthetic data for the private features, which preserves privacy by avoiding the use of sensitive features directly.
    \item \texttt{SGD\_pub}: A non-DP model trained on a public dataset, where sensitive features have been masked to gaussian noise. The model is update with:
    \[
    g = \nabla \mathcal{L}(x_{\text{pub}}, y).
    \]
    where \( x_{\text{pub}} \) is the public data.
    \item \texttt{DP-SGD} \cite{dpsgd}: A standard DP-SGD model that enforces sample-level differential privacy, protecting all features through clipping and noise addition during training. The model is update with:
    \[
    g = \text{DP-SGD}(\nabla \mathcal{L}\big(x_{\text{ori}}, y\big)) 
    \]
    \item \texttt{Na\"ive Fusion}: A naive baseline that combines gradients from \texttt{DP-SGD} (on original data) and \texttt{SGD\_hybrid} (on hybrid data). At each step, the model is updated with a weighted average of the two gradients:
    \[
    g = \lambda \cdot \text{DP-SGD}(\nabla \mathcal{L}\big(x_{\text{ori}}, y\big)) + (1 - \lambda) \cdot \nabla \mathcal{L}\big(x_{\text{hybrid}}, y\big),
    \]
    where \( \lambda \in [0, 1] \) controls the trade-off between the unbiased but noisy \texttt{DP-SGD} gradient and the noiseless but potentially biased \texttt{SGD\_hybrid} gradient. The sensitive features remain protected due to the application of DP-SGD on the original data. 
    \item \texttt{Na\"ive Fusion}: A naive baseline that combines gradients from \texttt{DP-SGD} (on original data) and \texttt{SGD\_pub} (on public data). At each step, the model is updated with a weighted average of the two gradients:
    \[
    g = \lambda \cdot \text{DP-SGD}(\nabla \mathcal{L}\big(x_{\text{ori}}, y\big)) + (1 - \lambda) \cdot \nabla \mathcal{L}\big(x_{\text{Pub}}, y\big),
    \]
    where $x_{\text{pub}}$ is the public data
    
    \item \texttt{Feature DP}: The algorithm used in \cite{featuredp}
    \item \texttt{Calibrated Fusion}: An ablation of \ours~ that removes the representation consistency regularization. This variant isolates the impact of the representation consistency loss in our framework, allowing us to evaluate how much the consistency signal contributes to overall performance and stability.
    \item \texttt{\ours}: Our proposed method.
\end{enumerate}

\subsection{All Hyperparameters}

\begin{table}[htbp]
\centering
\resizebox{\columnwidth}{!}{%
\begin{tabular}{|l|c|c|c|c|c|}
\hline
\textbf{Method} & \textbf{$\epsilon$} & \textbf{Epochs} & \textbf{$C$} & \textbf{$\alpha$} & \textbf{$\beta$} \\
\hline
FusionDP & 0.1 & 13 & 0.1 & 5.0 & 0.2 \\
FusionDP & 0.3 & 7 & 0.2 & 8.0 & 0.5 \\
FusionDP & 0.5 & 7 & 0.5 & 5.0 & 0.2 \\
FusionDP & 0.7 & 7 & 0.4 & 10.0 & 0.2 \\
FusionDP & 1.0 & 7 & 0.6 & 8.0 & 0.2 \\
FusionDP & 1.5 & 13 & 1.3 & 3.0 & 0.5 \\
FusionDP & 2.0 & 13 & 1.8 & 3.0 & 0.5 \\
DP-SGD & 0.1 & 7 & 0.2 & 0.0 & 0.0 \\
DP-SGD & 0.3 & 13 & 0.5 & 0.0 & 0.0 \\
DP-SGD & 0.5 & 13 & 1.0 & 0.0 & 0.0 \\
DP-SGD & 0.7 & 13 & 1.5 & 0.0 & 0.0 \\
DP-SGD & 1.0 & 13 & 2.5 & 0.0 & 0.0 \\
DP-SGD & 1.5 & 13 & 2.5 & 0.0 & 0.0 \\
DP-SGD & 2.0 & 13 & 2.5 & 0.0 & 0.0 \\
Feature-DP & 0.1 & 13 & 0.05 & 8.0 & 0.0 \\
Feature-DP & 0.3 & 13 & 0.3 & 3.0 & 0.0 \\
Feature-DP & 0.5 & 7 & 1.0 & 8.0 & 0.0 \\
Feature-DP & 0.7 & 13 & 1.5 & 5.0 & 0.0 \\
Feature-DP & 1.0 & 13 & 1.8 & 3.0 & 0.0 \\
Feature-DP & 1.5 & 7 & 2.0 & 5.0 & 0.0 \\
Feature-DP & 2.0 & 13 & 2.5 & 3.0 & 0.0 \\
Calibrated Fusion & 0.1 & 13 & 0.05 & 8.0 & 0.0 \\
Calibrated Fusion & 0.3 & 13 & 0.1 & 8.0 & 0.0 \\
Calibrated Fusion & 0.5 & 13 & 0.3 & 5.0 & 0.0 \\
Calibrated Fusion & 0.7 & 13 & 0.5 & 3.0 & 0.0 \\
Calibrated Fusion & 1.0 & 13 & 1.0 & 3.0 & 0.0 \\
Calibrated Fusion & 1.5 & 13 & 1.2 & 3.0 & 0.0 \\
Calibrated Fusion & 2.0 & 13 & 1.0 & 5.0 & 0.0 \\
Naive Fusion & 0.1 & 13 & 0.1 & 0.5 & 0.0 \\
Naive Fusion & 0.3 & 13 & 0.5 & 0.8 & 0.0 \\
Naive Fusion & 0.5 & 13 & 1.5 & 0.8 & 0.0 \\
Naive Fusion & 0.7 & 13 & 2.0 & 0.8 & 0.0 \\
Naive Fusion & 1.0 & 13 & 2.2 & 0.8 & 0.0 \\
Naive Fusion & 1.5 & 13 & 2.5 & 1.0 & 0.0 \\
Naive Fusion & 2.0 & 13 & 3.0 & 1.0 & 0.0 \\
Naive Fusion-Pub & 0.1 & 13 & 0.1 & 0.5 & 0.0 \\
Naive Fusion-Pub & 0.3 & 13 & 0.5 & 0.8 & 0.0 \\
Naive Fusion-Pub & 0.5 & 13 & 1.5 & 0.8 & 0.0 \\
Naive Fusion-Pub & 0.7 & 13 & 2.0 & 0.8 & 0.0 \\
Naive Fusion-Pub & 1.0 & 13 & 2.2 & 0.8 & 0.0 \\
Naive Fusion-Pub & 1.5 & 13 & 2.5 & 1.0 & 0.0 \\
Naive Fusion-Pub & 2.0 & 13 & 3.0 & 1.0 & 0.0 \\
\hline
\end{tabular}%
}
\caption{Hyperparameters for Sepsis prediction}
\label{tab:hyperparameters_sepsis}
\end{table}

\begin{table}[htbp]
\centering
\resizebox{\columnwidth}{!}{%
\begin{tabular}{|l|c|c|c|c|c|}
\hline
\textbf{Method} & \textbf{$\epsilon$} & \textbf{Epochs} & \textbf{$C$} & \textbf{$\alpha$} & \textbf{$\beta$} \\
\hline
FusionDP & 5.0 & 25 & 1.0 & 0.5 & 0.5 \\
DP-SGD & 5.0 & 30 & 1.5 & 0.0 & 0.0 \\
Feature-DP & 5.0 & 30 & 1.0 & 0.2 & 0.0 \\
Calibrated Fusion & 5.0 & 13 & 0.05 & 8.0 & 0.0 \\
Naive Fusion & 5.0 & 30 & 1.5 & 0.5 & 0.0 \\
Naive Fusion-Pub & 5.0 & 30 & 1.5 & 0.5 & 0.0 \\
\hline
\end{tabular}%
}
\caption{Hyperparameters for MIMIC-III clinical notes classification}
\label{tab:hyperparameters_mimic3}
\end{table}

\subsubsection{Hyperparameters Analysis.}

For tabular Sepsis prediction, we conduct a grid search over key hyperparameters: clipping norm $C \in [0.05, 3.0]$, loss-balancing coefficient $\alpha \in [0, 13]$, and representation consistency weight $\beta \in [0, 1]$ across privacy budgets $\epsilon$. The optimal parameters for each epsilon each method is listed in table \ref{tab:hyperparameters_sepsis}

For low privacy budgets ($\epsilon \leq 0.7$), we use smaller $C$ to control noise magnitude but compensate with larger $\alpha$ values to amplify the private loss contribution. At higher $\epsilon$, larger $C$ allows more informative gradients from the private branch, requiring smaller $\alpha$.

The consistency weight $\beta$ creates a fundamental trade-off: larger values improve representation alignment but increase gradient sensitivity and noise under strong privacy constraints. This can stabilize training and reduce imputation bias but may degrade utility without sufficient clipping norm.

Since these parameters interact, joint tuning of $\alpha$, $\beta$, and $C$ relative to $\epsilon$ is essential for optimal privacy-utility trade-offs.

Moreover, we observe that adding the representation consistency loss into the private branch in \ours increases sensitivity, as indicated by the higher optimal clipping norm $C$ compared to Calibrated Fusion. This requires injecting more noise to achieve the same privacy budget $\epsilon$. However, the increase remains moderate—\ours consistently uses smaller $C$ values than DP-SGD.  This suggests that representation alignment introduces additional sensitivity, but without incurring a substantial privacy cost, leading to a favorable trade-off that improves performance.

\subsection{Computing Infrastructure}

All experiments are conducted in Python on an H100 GPU with 80GB memory running a Linux-based operating system. The names and versions of all libraries used are provided in the code appendix.

\section{Tabular Classification Additional Details}
\label{app:tabular}

\textbf{Preprocessing Details.}
We preprocess the Sepsis dataset by retaining the most recent measurement for time-series features and discarding features with $>$70\% missing values. Given the severe class imbalance (7\% positive cases), we downsample the majority class to achieve $\sim$15\% positive examples. This preprocessing step helps mitigate the effects of label imbalance and allows for a clearer assessment of our method's effectiveness.

\textbf{Backbone Details.}
The MLP consists of four fully connected layers. The first three layers are each followed by a GELU activation, LayerNorm for stabilization, and dropout with a probability of 0.15 for regularization. The final layer outputs a single logit, suitable for binary classification. We omit any activation function at the output layer to allow direct use of the \texttt{BCEWithLogitsLoss} objective during training.

\section{Clinical Text Classification Additional Details}
\label{app:text}

Figure \ref{fig:hybrid_prompt} illustrates the LLM prompt for imputing the masked clinical text.
\begin{figure}[h]
\centering
\includegraphics[width=0.8\columnwidth]{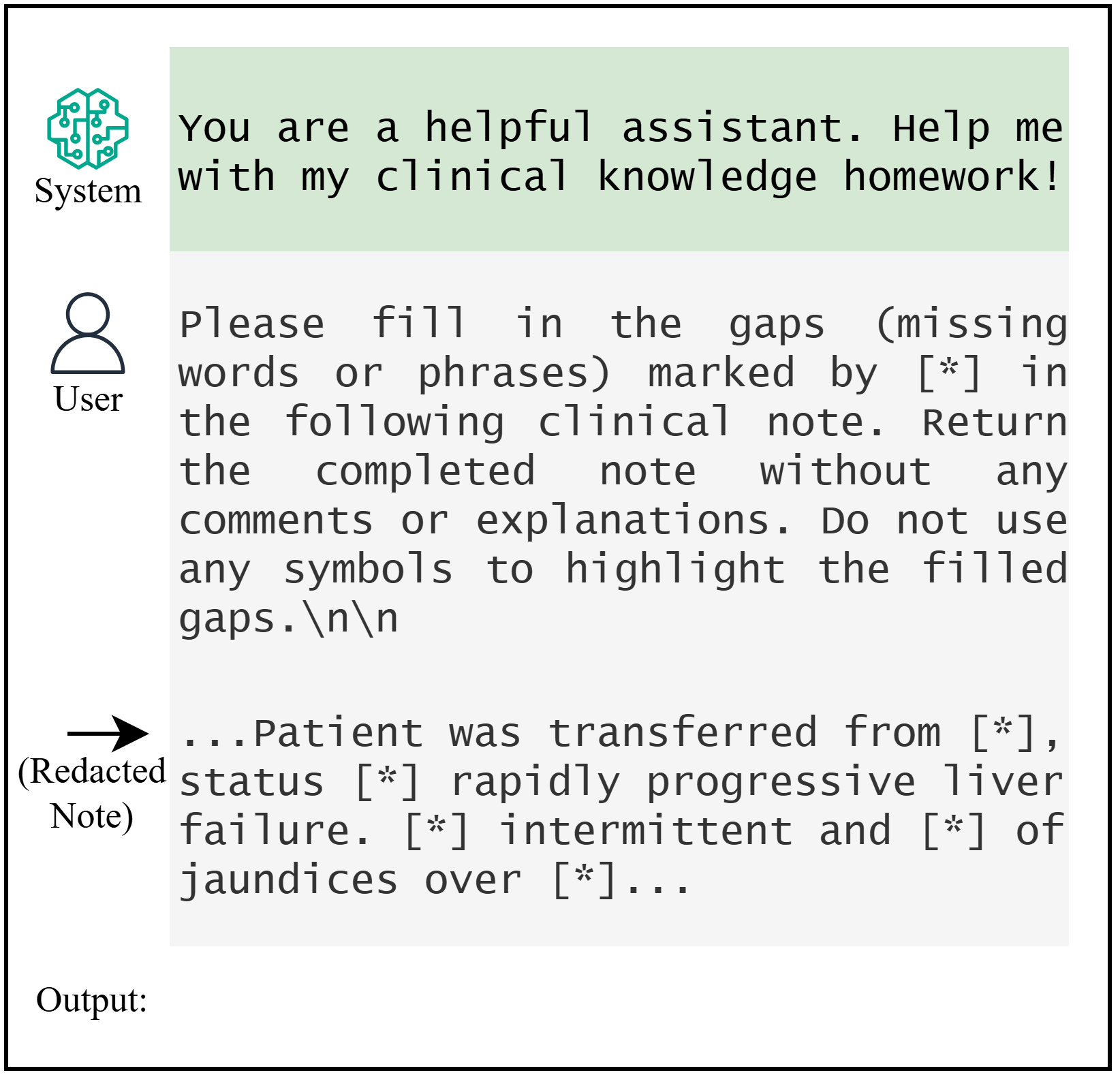} 
\caption{Prompt for imputing redacted notes with GPT-4o-mini}
\label{fig:hybrid_prompt}
\end{figure}

\textbf{Backbone.}
Compared to LAAT \cite{laat}, which employs a BiLSTM encoder and more sophisticated attention mechanisms, CAML’s lightweight, convolution-based architecture offers greater computational efficiency and more stable gradient behavior under DP-SGD, making it better suited for privacy-preserving training. As a result, we choose CAML as the backbone model.

Each clinical note is encoded using a one-dimensional convolutional layer applied to the word embeddings, followed by a ReLU activation. For each ICD-9 code, a learnable label query vector attends to the hidden representations via dot-product attention, producing a label-specific summary. Final predictions are obtained through dot-product scoring between the summary and the query vector, followed by a bias term.

\subsection{Measures of Variation}

The standard deviation across 5 runs for each method on MIMIC-III clinical notes classification are reported in Table \ref{tab:mimic3_std}.

\begin{table*}[t]
\centering
\footnotesize
\begin{tabular}{l cc c cc c cc c cc}
\hline
 & 
\multicolumn{2}{c}{\textbf{F1}} & &
\multicolumn{2}{c}{\textbf{Precision}} &&
\multicolumn{2}{c}{\textbf{Recall}} &&
\multicolumn{2}{c}{\textbf{AUC}} \\
\cline{2-3} \cline{5-6} \cline{8-9} \cline{11-12}
\textbf{Model} & Micro & Macro && Micro & Macro && Micro & Macro && Micro & Macro \\
\hline
SGD-ORG (Non-Private) & 0.01906 & 0.01142 && 0.018825& 0.02366 && 0.02587 & 0.01714 && 0.00709 & 0.01359 \\
DP-SGD (Sample-level DP) & 0.01413 & 0.00547 && 0.02991 & 0.01392 && 0.01474 & 0.01231 && 0.01587 & 0.01428 \\
\hline
SGD-Pub & 0.01499 & 0.01927 &&	0.01533 &	0.00864	&& 0.11355 &	0.11598 &&	0.02744	& 0.01714 \\

SGD-Hybrid & 0.01293 &	0.01441	&& 0.01923	& 0.00706 &&	0.04165	& 0.04987 &&	0.01494	& 0.00977 \\
\hline
Na\"ive Fusion-Pub & 0.00662	& 0.00461	&& 0.00905	& 0.00553	&& 0.00793	& 0.00612	&& 0.00749	& 0.01080 \\
Feature-DP & 0.00870	& 0.00482	&& 0.02115 &	0.01188	&& 0.00485	& 0.00325	&& 0.00854	& 0.00410 \\
\hline
Na\"ive Fusion & 0.00854	& 0.00557 &&	0.03374	& 0.00999	&& 0.01470	& 0.01339	&& 0.01287	& 0.01470 \\
Calibrated Fusion & 0.02221	& 0.01590	&& 0.01744	& 0.01584	&& 0.02236	& 0.01602	&& 0.00674	& 0.00875 \\
\ours & 0.00912	& 0.01382	&& 0.02215	& 0.02104	&& 0.00901	& 0.01031 &&	0.00688 &	0.00803 \\
\hline
\end{tabular}
\caption{Standard deviation across 5 runs for each method on MIMIC-III clinical notes classification with $\epsilon = 5.0$ and $\delta = N^{-1.1}$.}
\label{tab:mimic3_std}
\end{table*}

\subsection{Wilcoxon Signed-rank Test}

\begin{table*}[ht]
\centering
\small
\begin{tabular}{l cc c cc c cc c cc}
\hline
 & 
\multicolumn{2}{c}{\textbf{F1}} & &
\multicolumn{2}{c}{\textbf{Precision}} && 
\multicolumn{2}{c}{\textbf{Recall}} && 
\multicolumn{2}{c}{\textbf{AUC}} \\
\cline{2-3} \cline{5-6} \cline{8-9} \cline{11-12}
\textbf{Model} & Micro & Macro && Micro & Macro && Micro & Macro && Micro & Macro \\
\hline
Calibrated Fusion     & \textbf{0.0312} & \textbf{0.0312} && 0.0625 & \textbf{0.0312} && 0.8438 & 0.7812 && \textbf{0.0312} & \textbf{0.0312} \\
DP-SGD                & \textbf{0.0312} & \textbf{0.0312} && \textbf{0.0312} & \textbf{0.0312} && \textbf{0.0312} & 0.0625 && \textbf{0.0312} & \textbf{0.0312} \\
Feature-DP            & \textbf{0.0312} & \textbf{0.0312} && \textbf{0.0312} & \textbf{0.0312} && \textbf{0.0312} & \textbf{0.0312} && \textbf{0.0312} & \textbf{0.0312} \\
Naive Fusion          & \textbf{0.0312} & \textbf{0.0312} && 0.0625 & \textbf{0.0312} && 0.0625 & \textbf{0.0312} && \textbf{0.0312} & 0.0625 \\
Naive Fusion (Public) & \textbf{0.0312} & \textbf{0.0312} && \textbf{0.0312} & \textbf{0.0312} && \textbf{0.0312} & \textbf{0.0312} && \textbf{0.0312} & \textbf{0.0312} \\
\hline
\end{tabular}
\caption{Wilcoxon signed-rank test $p$-values comparing FusionDP to each baseline across evaluation metrics. Bold values ($p < 0.05$) indicate statistical significance.}
\label{tab:wilcoxon_results}
\end{table*}

Table \ref{tab:wilcoxon_results} presents one-sided Wilcoxon signed-rank test results comparing \ours~ against five private baselines across eight evaluation metrics on the MIMIC-III clinical text classification task, computed over 5 independent runs. Given the non-parametric nature of the Wilcoxon test and our small sample size ($n=5$, the smallest possible non-zero p-value is 0.0312—indicating that \ours~ outperformed the baseline in all 5 runs for that metric.

\ours~ consistently outperforms existing methods on core classification metrics (F1 and AUC), with statistically significant improvements across all runs ($p=0.0312$). While \ours~ maintains competitive recall performance, its gains in recall-micro and recall-macro over Calibrated Fusion are not statistically significant, suggesting that it preserves sensitivity while achieving substantial improvements in precision and ranking performance—particularly valuable in imbalanced classification settings. These results demonstrate that \ours~ delivers consistent and robust performance gains over prior methods.

\section{Additional Experiment for Sepsis Prediction}

\begin{figure}[h]
\centering
\includegraphics[width=1\columnwidth]{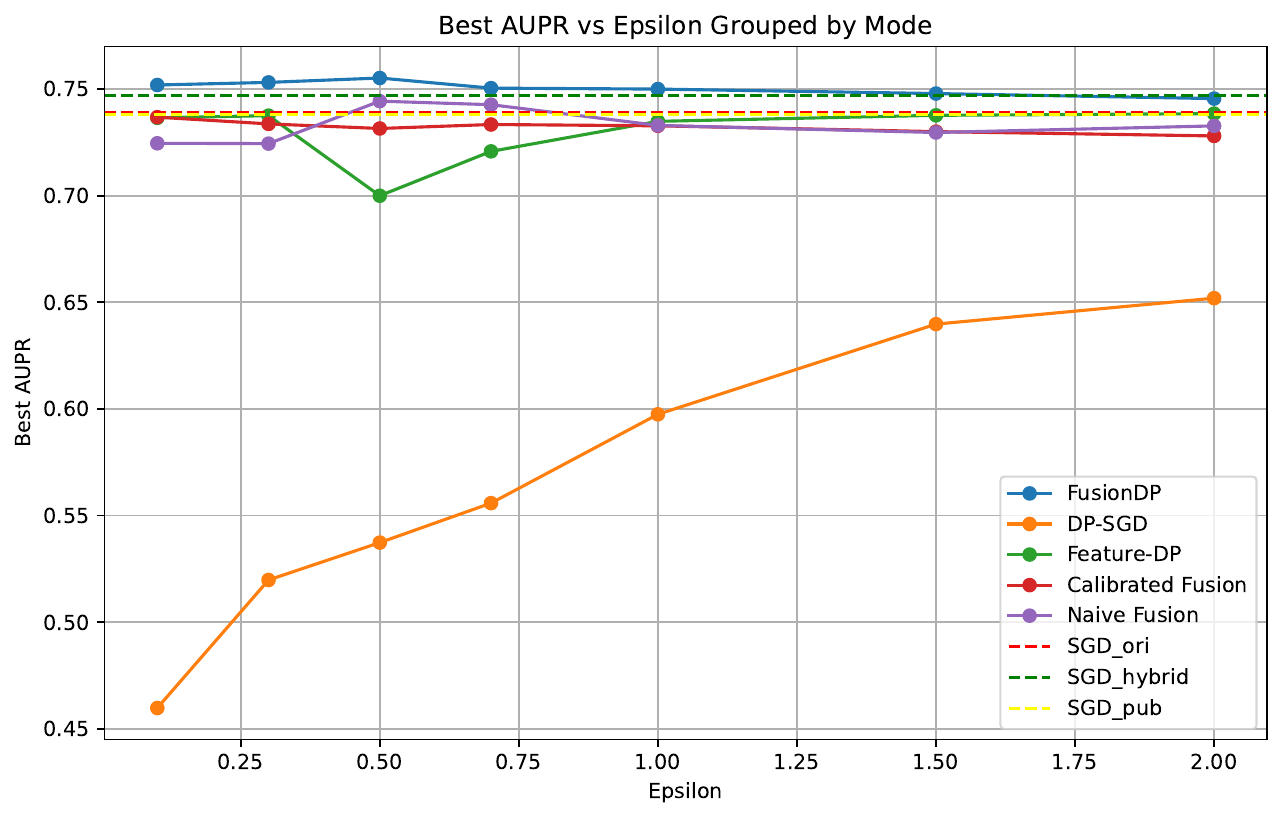} 
\caption{Experiment on Sepsis prediction with only age, gender, ICU unit type as sensitive features. \ours~ consistently outperforms other private baselines.}
\label{fig:hybrid_prompt}
\end{figure}

We conduct an additional experiment on sepsis prediction where we vary the selection of sensitive features. Specifically, we treat only age, gender, and ICU unit type as private, while considering the hours between hospital and ICU admission, ICU length of stay, all vital signs and laboratory values as public. In this setting, we observe that these three private features contribute relatively little to sepsis prediction—evidenced by the near-identical performance of SGD\_ori and SGD\_pub, and even slightly better performance from SGD\_hybrid. Notably, \ours maintains stable performance across all privacy budgets $\epsilon$, which may be attributed to the dominance of the public gradient (computed on hybrid data) in guiding convergence. Among all baselines with privacy guarantees, \ours consistently outperforms the others, further demonstrating its effectiveness even when the private features are weak predictors.


\end{document}